\documentclass[letterpaper, 10 pt, conference]{ieeeconf}  

\IEEEoverridecommandlockouts                              
\usepackage[utf8]{inputenc}
\usepackage[T1]{fontenc}
\usepackage{color}
\usepackage{soul}
\usepackage[table,usenames,dvipsnames]{xcolor} 
\usepackage{graphicx}
\usepackage{mathtools}
\usepackage{cases}

\usepackage{amsmath} 
\usepackage{amssymb}  
\usepackage{hyperref}

\usepackage{array}

\newtheorem{remark}{Remark}

\newcommand{\norm}[1]{\left\lVert#1\right\rVert}

\title{\LARGE \bf
Robust Trajectory Optimization over Uncertain Terrain with Stochastic Complementarity
}

\author{Luke Drnach$^{1}$ and Ye Zhao$^{2}$
\thanks{$^{1}$L. Drnach is with the School of Electrical and Computer Engineering,  Institute for Robotics and Intelligent Machines, Georgia Institute of Technology, Atlanta, GA, USA.
        {\tt\small luke.drnach@gatech.edu}}%
\thanks{$^{2}$Y. Zhao is with the G. W. Woodruff School of Mechanical Engineering at Georgia Institute of Technology, Atlanta, GA, USA.
        {\tt\small ye.zhao@me.gatech.edu}}%
\thanks{The authors are with the Laboratory for Intelligent Decision and Autonomous Robots, G. W. Woodruff School of Mechanical Engineering, Georgia Institute of Technology, Atlanta, GA, USA.}%
\thanks{A video of the paper is available at \url{ https://youtu.be/IdHXPEV1iRw}}
\thanks{*This work is supported by the National Science Foundation under grant Nos  DGE-1650044}
}
\begin{document}
\maketitle
\thispagestyle{empty}
\pagestyle{empty}
\begin{abstract}
Trajectory optimization with contact-rich behaviors has recently gained attention for generating diverse locomotion behaviors without pre-specified ground contact sequences. However, these approaches rely on precise models of robot dynamics and the terrain and are susceptible to uncertainty. Recent works have attempted to handle uncertainties in the system model, but few have investigated uncertainty in contact dynamics. In this study, we model uncertainty stemming from the terrain and design corresponding risk-sensitive objectives under the framework of contact-implicit trajectory optimization. In particular, we parameterize uncertainties from the terrain contact distance and friction coefficients using probability distributions and propose a corresponding expected residual minimization cost design approach. We evaluate our method in three simple robotic examples, including a legged hopping robot, and we benchmark one of our examples in simulation against a robust worst-case solution. We show that our risk-sensitive method produces contact-averse trajectories that are robust to terrain perturbations. Moreover, we demonstrate that the resulting trajectories converge to those generated by a traditional, non-robust method as the terrain model becomes more certain. Our study marks an important step towards a fully robust, contact-implicit approach suitable for deploying robots on real-world terrain.

\end{abstract}


\section{INTRODUCTION}

Trajectory optimization has become a powerful tool for designing dynamic motions for robots with nonlinear, hybrid, under-actuated dynamics and constraints~\cite{Ratliff09,Schulman14,mordatch2015ensemble,posa2014direct, Posa16,Kuindersma16}. Although impressive locomotion applications abound in the literature, these algorithms are far from real-world deployment: success depends critically on multiple factors including model fidelity, environmental uncertainty, and the ability to design effective closed-loop strategies for executing planned motions~\cite{mordatch2015ensemble, Posa16}. As optimal strategies often lie on the boundary of the feasible region, errors in the dynamic model could result in the planned trajectory becoming dynamically infeasible. Additionally, unmodeled disturbances from the environment can introduce deviations from the nominal trajectory, which propagate through the dynamics and can result in large errors over time. While fast online re-planning and robust low-level control can aid in recovering from local disturbances, incorporating and reasoning about robustness in the high-level planning could fundamentally improve the overall system performance.

Within trajectory optimization, designing reliable behaviors for dynamic robot locomotion tasks that require intermittent frictional contact has been a persistent challenge over the past few decades. Contact sequences and forces can be calculated during trajectory planning using contact-implicit trajectory optimization \cite{posa2014direct}; however, this method requires exact knowledge of the terrain geometry and friction coefficients beforehand. As friction coefficients require specialized sensors to  estimate and real-world terrain geometry can be intractable to model outside the laboratory, the contact-implicit method becomes highly prone to errors and failures. For instance, errors in modeling the friction characteristics of a terrain could cause a robot to slip, and errors in modeling the geometry of the terrain could cause the robot to trip, both of which could result in a fall.  We hypothesize that the failure to explicitly account for uncertainties and feedback during trajectory design is a key contributor to slow progress in translating trajectory optimization, and in particular contact-implicit trajectory optimization, research results into a depolyable technology. 

Our study takes one step toward addressing this problem by deriving a risk-sensitive variant of contact-implicit trajectory optimization. We develop objective functions derived from statistics related to the traditional complementarity constraints for contact and reason about the robustness by comparing trajectories generated by our robust method to those generated using the conventional complementarity constraint method. To contribute specifically to the field, we:
\begin{itemize}
    \item Include parametric models of uncertainty in the friction coefficient and in the contact distance into contact-implicit trajectory optimization.
    \item Develop risk-sensitive objectives that produce contact-averse trajectories which are robust to perturbations in terrain parameters when the terrain model is uncertain. 
    \item Demonstrate that our contact-sensitive method represents a smooth generalization of the traditional complementarity constraints in that our method converges to the complementarity method as the contact parameters become certain. 
\end{itemize}
We evaluate our framework in three examples and benchmark one of our robotic examples against a worst-case robust approach in simulation.  We show that the control trajectories resulting from our optimization approach are robust to perturbations in the contact parameters, since uncertainty in the contact constraints is explicitly modeled. Although our work assumes the contact constraints are uncertain, and thus we cannot enforce the exact physical constraints, we show that the solution sets of our robust objective correspond with the solution of the complementarity constraints, and we prove this correspondence in limiting cases. Thus, trajectories generated under our robust objective may facilitate implementing robust motion plans on physical robots.


\section{Related Work}

\subsection{Contact-Implicit Trajectory Optimization}

Contact-implicit trajectory optimization, as pioneered by the work in \cite{posa2014direct,manchester_contact-implicit_2019, patel_contact-implicit_2019, Mordatch12, sleiman2019contact, dai2014whole}, includes contact forces as decision variables in an optimal control problem. The contact forces are governed set of complementarity constraints \cite{Stewart96}, and the resulting direct transcription problem is solved through a large-scale nonlinear program such as sequential quadratic programming (SQP). 
Compared to approaches with predefined contact sequence \cite{pardo2017hybrid, carpentier2017learning},  a remarkable advantage of this contact-implicit method lies in avoiding exhaustive search of combinatorial contact mode possibilities, which are computationally prohibitive for contact-rich robotic systems. Building on top of this contact-implicit approach, our study has a large focus on reasoning about robustness to uncertainties with respect to contact surface geometry and friction properties. 

\subsection{Robust Trajectory Optimization}
Reasoning about the robustness of trajectory optimization has been extensively explored in robotics \cite{luo2017robust, johnson2016convergent, pan2015robust}. One well-received robust approach is ensemble contact-invariant optimization \cite{mordatch2015ensemble} which samples uncertain physical model parameters and generates a collection of specific model instances. Trajectories associated with each model instance are coupled via a penalty cost and a single nominal trajectory is generated with a notion of robustness. 
In more recent works \cite{luo2017robust, song_identifying_2020}, uncertainty in friction coefficients has been addressed by updating model parameters from errors between planned motions and simulated or experimental motions; however, these methods require multiple physical interactions to improve the estimate of the friction coefficient, and early interactions can fail due to a lack of robustness. Differing from modeling model parameter uncertainties or learning friction parameters, our study reasons about robustness to contact uncertainties, which is critical for safe contact-rich planning.

Risk-sensitive optimal control, a powerful approach to reason about robustness, employs high-order statistics in the cost function design \cite{kuindersma2013variable, singh2018framework, farshidian2015risk, ponton2016risk}. A seminal work in \cite{jacobson1973optimal} proposed a risk-sensitive Linear-Exponential Gaussian algorithm which includes the high-order statistics by using the expectation of the exponential transformation of a performance index as the cost. In these risk-sensitive works, uncertainty is assumed to enter through either the estimation of the states or through the actuation of controls, and the cost function is transformed to produce risk-sensitive behaviors. However, these works have yet to address uncertainty from constraints dealing with contact. Here we consider uncertainty arising from the contact model, which is normally included in trajectory optimization as complementarity constraints, and we derive additional cost terms to produce risk-sensitive behaviors. 


\subsection{Stochastic Complementarity Problems}
One approach to handling uncertainty in complementarity constraints is to recast them as expected residual minimization (ERM) problems. The ERM formulation, which is a smooth alternative for both linear and nonlinear complementarity constraints (LCPs and NCPs), has been extensively investigated in the context of stochastic complementarity problems (SCPs) \cite{chen2005expected}. Smoothed residual functions are often introduced as approximations of the original  constraints \cite{chen1996class}, and solutions to the ERM problem are robust in the sense that they have minimum sensitivity to random SCP parameter variations. Another approach is to cast the complementarity problem as a worst-case robust optimization, as in \cite{xie2016robust}. In the case of LCPs, the worst-case variant can be solved by a single convex program. 
Nevertheless, the convexity assumption is conservative since many robotic problems are inherently non-convex and nonlinear. Moreover, application of both the ERM and the worst-case methods to trajectory optimization has been largely under-explored.
An initial effort applied the ERM framework to solve robotic problems with stochastic complementarity \cite{tassastochastic}, where uncertainty is assumed to be derived from errors in state estimation. However, that work mainly applies the ERM method as a smoothing technique so the complementarity constraints could be included in indirect trajectory optimization. In our study, we further explore the ERM technique as a method for encoding uncertainty about the terrain model into direct trajectory optimization and explicitly analyze the robustness of the resulting trajectories.

\section{PROBLEM FORMULATION}
\subsection{Contact-Implicit Trajectory Optimization}
The goal of contact-implicit trajectory optimization for rigid bodies with intermittent contact is to find the states $x$, controls $u$, and contact forces $\lambda$ that solve the following optimal control problem:
\begin{subequations}
    \begin{align} \label{eq:OptimalControl}
        \min_{x, u, \lambda} \int_0^{T}L(x, u, \lambda)dt + L_F(x(T))
    \end{align}
    \begin{numcases}{\text{s.t.}} \label{eq:Dynamics}
        M(q)\ddot{q} + C(\dot{q},q) = B u + J_c^\top(q)\lambda \\ \nonumber
        x(0) = x_0, x(T) = x_f \\ \label{eq:DistanceConstraint}
        0 \leq \lambda_N \perp \phi(q) \geq 0 \\ \label{eq:SlidingConstraint}
        0 \leq \lambda_T \perp \gamma + J_T\dot{q}  \geq 0 \\ \label{eq:FrictionConstraint}
        0 \leq \gamma \perp \mu\lambda_N - e^\top\lambda_T \geq 0  
    \end{numcases}
\end{subequations}
where $x = (q, \dot{q})$ is the state, $q$ is the system configuration, $x_0$ and $x_f$ are the initial and final states respectively, $L$ and $L_F$ are the running and terminal costs respectively, Eq. \eqref{eq:Dynamics} represents the rigid-body dynamics with mass matrix $M$, Coriolis and conservative forces $C$, control selection matrix $B$, and contact Jacobian $J_c$. Eqs. \eqref{eq:DistanceConstraint}-\eqref{eq:FrictionConstraint} are the nonlinear complementarity constraints encoding the contact conditions. Eq. \eqref{eq:DistanceConstraint} encodes a normal distance constraint, where $\lambda_N$ is the normal force and $\phi(q)$ is the normal distance. Eq. \eqref{eq:SlidingConstraint} encodes a constraint on the sliding velocity, where $\gamma$ is a slack variable related to the magnitude of the sliding velocity, $J_T$ is the tangential part of the contact Jacobian, and $\lambda_T$ is the tangential contact force. Eq. \eqref{eq:FrictionConstraint} encodes a linearized friction cone constraint, where $\mu$ is the coefficient of friction and $e$ is a vector of 1s. The shorthand $ 0 \leq a \perp b \geq 0$ denotes a complementarity constraint: $a \geq 0, b\geq 0, a^\top b = 0$.

Numerical methods have already been developed to solve the problem \eqref{eq:OptimalControl}-\eqref{eq:FrictionConstraint} using either direct \cite{posa2014direct, manchester_contact-implicit_2019, patel_contact-implicit_2019} or indirect \cite{tassastochastic, carius_trajectory_2018} methods. In this work, instead of developing a more computationally efficient or more accurate high-order method, as was the case in \cite{manchester_contact-implicit_2019, carius_trajectory_2018}, and \cite{patel_contact-implicit_2019}, our goal is to develop and evaluate a framework for including contact uncertainties. Thus, we used a direct transcription method to convert the continuous dynamics and costs into their discrete analogs. We evaluated the dynamics using backward Euler integration and enforced the contact constraints at the end of each interval. Throughout our work, we used a quadratic cost on states and controls:
\begin{align*}
    L(x,u,\lambda) = \frac{1}{2}\left((x - x_f)^\top Q (x-x_f) + u^\top R u\right).
\end{align*}

\subsection{Stochastic Complementarity Constraints}
The preceding formulation assumes perfect knowledge of the contact parameters. If any of the terms in \eqref{eq:DistanceConstraint}-\eqref{eq:FrictionConstraint} are uncertain or random, then resolving the contact forces becomes a stochastic complementarity problem (SCP) \cite{luo_SCP_2013}:
\begin{align}
    &0 \leq z \perp F(z, \omega) \geq 0, \quad \omega\in\Omega \label{eq:SCP}
\end{align}
where $\omega$ represents a random quantity on probability space $(\Omega, \mathcal{F}, \mathcal{P})$ with given probability distribution $\mathcal{P}$, $z$ is the decision variable, and $F(\cdot)$ is a vector-valued function. Because $\omega$ is stochastic, the problem in \eqref{eq:SCP} is not well-defined and in general will not have a solution for all $\omega \in \Omega$. One approach, the expected value approach, is to replace the function $F$ with its expected value:
\begin{align}\label{eq:expected-residual}
    0 \leq z \perp \mathbb{E}[F(z, \omega)] \geq 0
\end{align}
The expected value method is largely equivalent to solving the deterministic problem at the mean value of $F$, and is not expected to be robust to random variations in the parameters.

\subsection{Expected Residual Minimization}
Theoretical works have studied robust solutions to Eq. \eqref{eq:SCP} in both the case when $F$ is affine \cite{chen2005expected, chen1996class, chen_robust_2009} and in the case when $F$ is nonlinear \cite{luo_SCP_2013}. In these works, it is common to define a residual function $\psi$ such that the residual is zero when the complementarity conditions are satisfied:
\begin{align}\label{eq:LCPequivalence}
    \psi(a, b) = 0 \Longleftrightarrow a \geq 0, b \geq 0, a^T b = 0
\end{align}
One common choices for the residual function is the "min" function $\psi_{\rm min}(a,b) = {\rm min}(a,b)$.
Then, the expectation of the residual can be taken to form a deterministic objective for the original SCP, which can then be minimized. This is the Expected Residual Minimization (ERM) approach that we use in this work, which is commonly formulated as:
\begin{align}\label{eq:ERM}
    \min_{z} \mathbb{E}[\norm{\psi(z, F(z, \omega))}^2]
\end{align}

The ERM formulation Eq~\eqref{eq:ERM} has an advantage over the expected value method, in that it has minimum sensitivity to variations in the random parameters \cite{chen_robust_2009}:
\begin{align*}
    \mathbb{E}[\norm{\psi}^2] = \norm{\mathbb{E}[\psi]}^2 + \mathbb{E}[\norm{\psi - \mathbb{E}[\psi]}^2] 
\end{align*}
where, for vector-valued $\psi$, $\mathbb{E}[\norm{\psi-\mathbb{E}[\psi]}^2] = \text{tr}(\text{Cov}(\psi))$ is the trace of the covariance matrix, or the total variance. Thus, the ERM approach minimizes the mean-squared residual and the total variation with respect to random parameters.

\section{Contact-Robust Trajectory Optimization}
\subsection{Stochastic Complementarity in Trajectory Optimization}
Previous work using SCPs in trajectory optimization developed ERM closed-form cost functions for the special case when the elements of $F$ are normally distributed or logistically distributed \cite{tassastochastic}. In this work, we make use of the ERM for Gaussian distributed variables:
\begin{gather}\nonumber
    F \sim \mathcal{N}(\mu_F, \sigma_F)\\ \nonumber
    \mathbb{E}[\min(z,F)^2] =\\ \label{eq:GaussERM}
    z^2 - \sigma_F^2(z+\mu_F)p(z) + (\sigma_F^2 + \mu_F^2 - z^2)P(z) \\ \nonumber
    p(z) = \frac{1}{\sigma_F\sqrt{2\pi}}e^{-\frac{1}{2}(\frac{z - \mu_F}{\sigma_F})^2} \\ \nonumber
    P(z) = \int_{-\infty}^zp(t)dt = \frac{1}{2}(1+\text{erf}(\frac{z - \mu_F}{\sigma_F\sqrt{2}}))
\end{gather}
where $p(z)$ and $P(z)$ are the probability density and cumulative density functions for the normal distribution, respectively, evaluated at $z$. However, the previous work assumed the uncertainty resulted from state estimation and propagated directly to the SCP function $F$ \cite{tassastochastic}, which may not be consistent with the true uncertainty effects. 

In this work, we assume the uncertainty lies directly in the contact parameters - specifically the friction coefficient $\mu$ and normal distance $\phi$ - and derive the corresponding distributions $F$. We replace contact constraints with an ERM cost which encodes uncertainty about the terrain parameters:
\begin{align}\label{eq:ERM2}
    \min_{\mathbf{x,u,\lambda}} \sum_{i=0}^{N-1} \left(L(x_i, u_i, \lambda_i) + \beta\mathbb{E}[\norm{\psi(z_i, F(z_i, \omega))}^2]\right)
\end{align}
 where $\beta$ is a weighting scalar for the ERM objective and $N$ represents the total number of knot points, $\mathbf{x, u}$ and $\mathbf{\lambda}$ represent collections of the respective variables across knot points, and $z_i\in\{\mathbf{x, \lambda}\}$ represents the variables in the complementarity constraints. Except where noted otherwise, we used $\beta = 10^5$. In our preliminary work (results not discussed in this paper), we found that this value of $\beta$ was necessary to keep the ERM cost on  the same order of magnitude as the other costs. 

\begin{remark}
A notable feature of the Gaussian ERM is that, as the uncertainty decreases, the ERM objective function approaches the residual function objective evaluated at the mean value of the uncertain variable:
\begin{align}\label{eq:ERMConvergence}
    \lim_{\sigma\rightarrow0^+} \mathbb{E}[\min(z, F)^2] = \min(z, \mu_F)^2
\end{align}
This property can be proved by using L'Hopital's rule to show that $\lim_{\sigma\rightarrow0^+}\sigma^2p(z) = 0$ and that, for the cumulative distribution function:
\begin{align*}
    \lim_{\sigma\rightarrow0^+}P(z) = 
    \begin{cases}
        1, \quad z - \mu_F > 0 \\
        0, \quad z - \mu_F < 0
    \end{cases}
\end{align*}
\end{remark}

\subsection{Worst-case Optimization}
To compare against the ERM formulation, we also consider a worst-case scenario of the LCP formulation as a robust optimization \cite{xie2016robust}.
A general robust counterpart (RC) of the uncertain optimization problem can be formulated as:
\begin{align}\label{eq:robustOptFormulation}
& \underset{z \geq 0}{\text{min}} \;
\underset{\omega \in \Omega}{\text{max}} \; \psi(z, F(z, \omega))\\
& {\rm s.t.} \quad \underset{\omega \in \Omega}{\text{min}} \; F_i(z, \omega) \geq 0, \forall i \in \mathcal{I}
\end{align}
where the index set $\mathcal{I}$ comprises all the possible LCP instances.
However, this RC is computationally challenging to solve in general. To derive a tractable RC, we reformulate the LCP mathematical program by posing an $\infty$-norm  assumption on the uncertainty set $\Omega$ as
$
\Omega_\infty = \{\omega: ||\omega||_\infty \leq 1\}.
$
Given the uncertainty set above, we can express the residual value, $F(z, \omega)$ as
\begin{align}\nonumber
F(z, \omega) = F_0(z) + \sum_{k=1}^K \omega_k F_k(z)
\end{align}
Accordingly, the robust optimization of LCP contact dynamics becomes
\begin{align}\nonumber
& \underset{z \geq 0, \epsilon}{\text{min}}
\quad \epsilon 
\\\label{eq:robustOptFormulation2}
& \text{s.t.}
\quad F_k(z) \geq 0, \;  z F_k(z) \leq \epsilon, \; \forall k \in \{1,\ldots, K\}
\end{align}
where $\epsilon$ is a slack variable to minimize. 
Compared with the ERM formulation, this robust LCP formulation does not require a probability distribution over the uncertain parameters. Different from the expected value formulation which minimizes the mean scenario, the robust formulation version reasons about the worst-case scenario by enumerating $K$ LCP instances in the uncertainty set $\Omega_\infty$. In practice, the worst-case scenario corresponds to a specific LCP instance; in this work we benchmark the performance of the ERM against this worst-case instance in one of our examples.

\subsection{Characterizing Physical Contact Uncertainties}
\label{subsec:uncertainty}
\begin{figure}
 \centering
\includegraphics[width=0.8\linewidth]{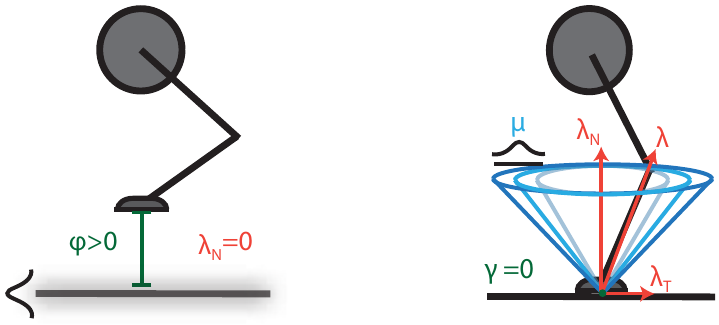}
 \caption{Contact geometry with terrain uncertainties from terrain height (left) and friction coefficient (right).}
 \label{fig:contact_geometry}
 \vspace{-5mm}
\end{figure}
Here we explicitly parameterize uncertainties in the friction cone and in the contact geometry and develop the corresponding ERM cost functions. Specifically, we assume a normally distributed friction coefficient and a normally distributed error in the distance to the terrain and then derive the corresponding distributions used in the ERM objective.

\subsubsection{Uncertainty in the friction coefficient}
We assume the friction coefficient $\mu$ is normally distributed with mean $\bar{\mu}$ and standard deviation $\sigma_\mu: \mu \sim \mathcal{N}(\bar{\mu}, \sigma_\mu)$. By linearity of the normal distribution, the friction cone defect $F_{FC}$ is also normally distributed:
\begin{align}\label{eq:FrictionERMDistribution}
    F_{FC} = \mu\lambda_N - e\lambda_T \sim \mathcal{N}(\bar{\mu}\lambda_N - e\lambda_T, \sigma_\mu\lambda_N)
\end{align}
Thus, we can replace the constraint \eqref{eq:FrictionConstraint} with the ERM objective \eqref{eq:GaussERM}, where $\mu_F = \bar{\mu}\lambda_N - e\lambda_T$ and $\sigma_F= \sigma_{\mu}\lambda_N$.

\subsubsection{Uncertainty in the contact distance} 
We assume the terrain is flat but that the contact distance is uncertain. The normal distance $\phi$ to the terrain is:
\begin{align*}
    \phi(q) = \eta^\top(p(q) - r)
\end{align*}
where $\eta$ is surface normal of the terrain, $p(q)$ is the Cartesian position of the robot end-effector, and $r$ is the position of the nearest contact point on the terrain surface. In this work we assume the normal distance $\phi$ is normally distributed: $\phi(q)\sim\mathcal{N}(\bar{\phi}(q), \sigma_\phi)$. In practice, the uncertainty can vary along the terrain and become a function of the robot configuration, i.e., $\sigma_\phi = \sigma_\phi(q)$. In either case, we can replace the normal distance constraint \eqref{eq:DistanceConstraint} with the ERM objective \eqref{eq:GaussERM}, where $\mu_F = \bar{\phi}(q)$ and $\sigma_F = \sigma_\phi$.
\begin{remark}
In theory, uncertainty in contact distance can also be expressed as uncertainty in the Cartesian coordinates of the nearest contact point, $r \sim \mathcal{N}(\bar{r}, \Sigma_r)$. However, because the terrain orientation $\eta$ is known, the normal distance becomes $\phi \sim \mathcal{N}(\eta^\top(p(q) - \bar{r}), \eta^\top \Sigma_R \eta)$, which is equivalent to the preceding formulation in terms of normal distance. 
\end{remark}
\begin{remark}
Our formulation for uncertainty in the contact distance can be reformulated to account for uncertainty in the terrain orientation $\eta$. If we assume the contact distance is known but that the terrain orientation is normally distributed - i.e. $\eta \sim \mathcal{N}(\bar{\eta}, \Sigma_\eta)$, then the normal distance follows a normal distribution: 
\begin{align*}
    \phi \sim \mathcal{N}\left(\bar{\eta}^\top(p(q)-r), (p(q)-r)^\top\Sigma_\eta (p(q)-r)\right)
\end{align*}
which is again equivalent to a distribution over the normal distance: $\phi\sim\mathcal{N}(\bar{\phi}(q), \sigma(q)_\phi)$. However, as the terrain orientation $\eta$ also partly defines the contact Jacobian $J_c$, additional care should be taken to ensure the uncertainty effects are consistent across the normal distance, the sliding velocity, and the dynamics constraints in Eqs. \eqref{eq:Dynamics}, \eqref{eq:DistanceConstraint}, and \eqref{eq:SlidingConstraint} respectively. Propagating uncertainty effects to the dynamics and deriving a corresponding risk-sensitive cost could be possible but is beyond the scope of the ERM framework we pursue here.

\end{remark}

\section{SIMULATION EXPERIMENTS}
Here we detail a set of simulation experiments to compare our ERM formulation to a baseline with non-stochastic nonlinear complementarity constraints, which we will refer to as the "non-robust" case. We study three examples: sliding a block over a surface with friction, a cart on frictionless rails that propels and stops itself through a pole in contact with the ground, and a single legged hopper with contact points at the toe and heel. We compare the trajectories generated by our ERM formulation to those generated by the non-robust case for a range of uncertainty parameter values. All of our trajectory optimization examples were implemented in MATLAB using Drake \cite{drake} and solved using SNOPT \cite{GilMS05}. Our code is available at \url{https://github.com/GTLIDAR/RobustContactERM}.

\subsection{Sliding a Block over a Surface with Unknown Friction} 
To benchmark the performance of the ERM method against the traditional non-robust trajectory optimization, we first study a two-dimensional 1kg block with height 1m sliding over a surface with uncertain friction (see Fig.~\ref{fig:block_simulation}(d)). The configuration of the block $q=[x_b, z_b]^\top$ is given by the planar CoM of the block and the control is a horizontal force acting on the block. The goal is to travel from the initial state, $x_0 = [0,0.5,0,0]^\top$, to the final state, $x_N = [5, 0.5, 0, 0]^\top$, in 1 second.  The running cost has weight matrices $R = 100I_2$ and $Q = I_4$, where $I_n$ is the $n\times n$ identity matrix. We used 101 knot points in the discretization, which corresponds to a timestep of $dt=0.01s$. 
The reference, non-robust trajectory was generated using the nonlinear complementarity constraints for contact and a friction coefficient of $\mu=0.5$. The ERM trajectories were generated using the ERM cost for friction with $\bar{\mu}=0.5$ as the mean value and 9 values of the standard deviation logarithmically spaced between $\sigma=0.001$ and $\sigma=1.0$. We compared the trajectories generated using the ERM cost to the nominal trajectory using a mean-squared difference criterion:
\begin{align*}
    \textnormal{MSD} = \frac{1}{N}\sum_{i=0}^{N-1}\norm{w_{\rm ERM}(t_i) - w_{\rm NCC}(t_i)}^2
\end{align*}
where $w_{\rm NCC}$ represents the state $x$, control $u$, or contact force $\lambda$ trajectory generated using the nonlinear complementarity constraint and $w_{\rm ERM}$ represents the corresponding trajectory generated using the ERM method. 

We compared the open-loop performance of the ERM controls to the non-robust controls in simulation using a time-stepping approach \cite{Stewart96, anitescu1997formulating}. All simulations started from the initial state $x_0$ and ran for 1 second. We compared the ERM simulations to the non-robust control simulation for 10 values of the terrain friction coefficient linearly spaced between $\mu=0.3$ and $\mu=0.7$. We also compared the simulations to a control generated using the worst-case scenario, where the friction uncertainty set was considered to be $\mu\in[0.3,0.7]$ - in this case, the worst-case solution corresponds to using the lowest friction coefficient value, $\mu=0.3$. We quantified the performance of the controls as the difference between target position of the block and the position achieved after 1s. 

\subsection{A Contact Driven Cart with Unknown Terrain Height}
Our second example is a double-pendulum connected to a cart, which is constrained to move horizontally but not vertically. In this example, the cart must use the two pendulums to contact the ground to push off and stop itself (see Fig.~\ref{fig:ContactCart}(a)). The mass of the cart and all the pendulums is $1$ kg, the pendulums each have length $1$ m, and the pendulum CoMs are halfway down their lengths. The configuration of the cart is $q = [x_c, \theta_1, \theta_2]$, where $x_c$ is the horizontal position of the center of the cart and $\theta_1$ and $\theta_2$ are the angles of the pendulum. The controls are the pendulum joint torques.

The goal is for the cart to travel from  $x_{c,0} = 0$m to $x_{c,N} = 5$m in 1s, starting and stopping at rest and with the pendulum end-effector in contact with the terrain. We used a quadratic cost with $R=\text{diag}([1,1])$ and $Q=\text{diag}([1, 100, 10, 1, 100, 10])$. We used 101 knot points, for a timestep of $dt=0.01s$. We encoded uncertainty about the terrain height in an ERM cost and compared the ERM trajectories against the reference, non-robust trajectory. We assumed a flat terrain with a mean distance of 1.5m from the center of the cart, and tested 7 values of height uncertainty logarithmically spaced between $\sigma = 0.001$ and $\sigma = 1$.
%
\subsection{A Single-Legged Hopper}
Our final example is similar to the contact-driven cart, except the vertical motion of the hopper is free and the hopper has a foot with contact points at the toe and heel (see Fig.~\ref{fig:HopperJump} (a)). Thus the configuration of the system is $q = [x_c, y_c, \theta_1, \theta_2, \theta_3]$ and the controls are the torques on the joints. Unlike the previous example, in which the cart only needed contact to start and stop, the single-legged hopper may require several steps to reach the goal position. In this example, the hopper must traverse 4m in 3 seconds, starting and stopping at rest. The weights in the running cost were $R=\text{diag}([0.01, 0.01, 0.01])$ and $Q=\text{diag}([1,10,10,100,100,1,1,1,1,1])$. We used 101 knot points in this example, corresponding to a fixed step size of $dt=0.03s$. We encoded uncertainty in both friction and in terrain height in an ERM cost  and compared the resulting trajectories to one generated without the uncertainties. In our experiments, we used a friction uncertainty of $\sigma=0.01$ and tested four uncertainties for the terrain height, $\sigma\in\{0.05, 0.09, 0.30, 0.50\}$. For both friction and height uncertainty, we weighted the ERM cost by a factor $\beta=10^4$.

\section{RESULTS}
\subsection{ERM Biases away from Contact Interaction}
\begin{figure}[t!]
    \centering
    \includegraphics[width=\linewidth]{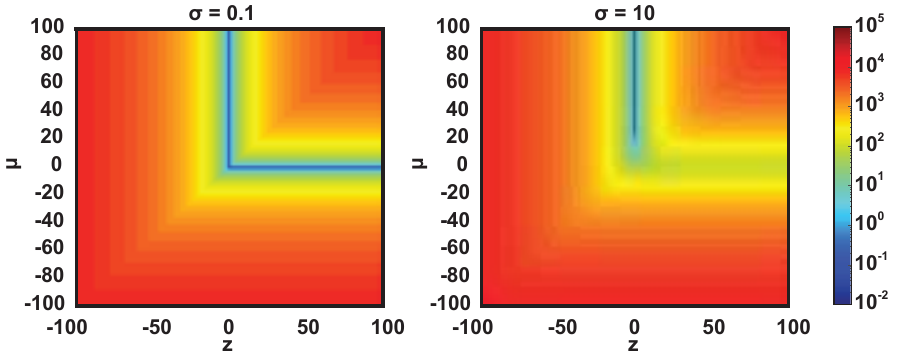}
    \caption{ERM cost map for different values of uncertainty. The horizontal axis represents the decision variable $z$ while the vertical axis represents the mean $\mu$ of the uncertain constraint. All subfigures share the same color axis, which represents the value of the ERM function. For uncertainty less than 1, the minima of the ERM cost approaches the nonnegative axes. As uncertainty increases, the values on the positive $\mu$-axis still represent local minima, but only at higher values of $\mu$.}
    \label{fig:ERMCostMap}
\end{figure}
To better understand the effect of nonzero uncertainty on the solutions, we mapped the ERM cost landscapes for different values of uncertainty (Figure \ref{fig:ERMCostMap}). At $\sigma = 0.1$, the ERM costmap has a set of low values near the nonnegative axes, which supports the claimed property that the ERM converges to the deterministic complementarity constraint when the uncertainty vanishes (Eq. \eqref{eq:ERMConvergence}). However, as the uncertainty increases, the cost along the decision variable axis increases and, when the uncertainty is high enough, the cost for low values of the mean of the uncertain constraint also increases (Figure \ref{fig:ERMCostMap}, $\sigma$ = 10). For contact problems, a high uncertainty should bias the optimal trajectory towards reducing the friction force (therefore increasing the friction cone residuals) for uncertain friction and towards increasing ground clearance for uncertain terrain height.

\subsection{ERM Generates Controls Robust to Changes in Friction}
\begin{figure}
    \centering
    \includegraphics[width=\linewidth]{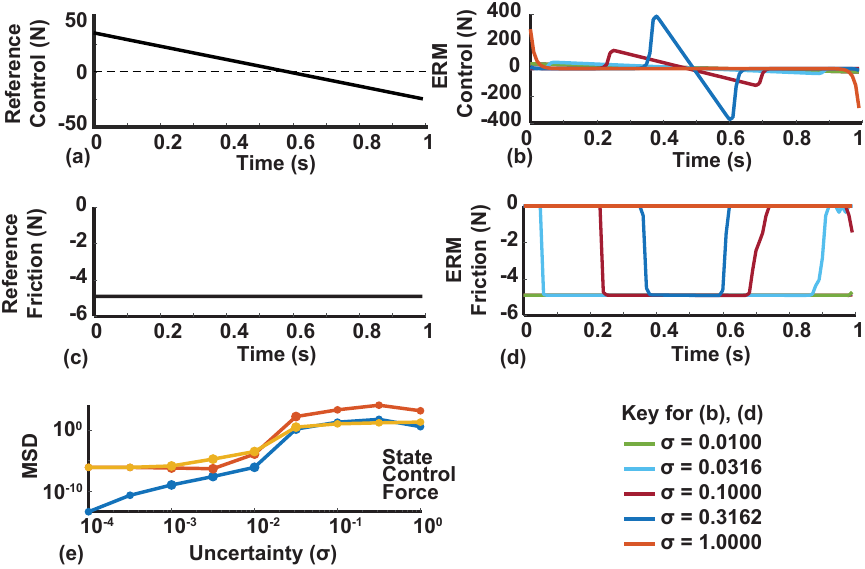}
    \caption{Trajectories generated by the ERM method at different levels of uncertainty compared to the reference trajectory, for the same value of the friction coefficient. The control force (a) and frictional force (c) in the reference, non-ERM trajectory are linear for the entire motion. Under difference values of uncertainty, the control (b) and friction (d) force change and become shorter in duration. (e) As the uncertainty decreases, the ERM trajectories converge to the reference trajectories. All trajectories were generated using an expected friction coefficient $\bar{\mu}=0.5$.}
    \label{fig:block_trajectories}
    \vspace{-5mm}
\end{figure}

For low values of uncertainty ($\sigma \leq 0.01$), the ERM method produced trajectories that were nearly indistinguishable from the reference trajectory generated by the nonlinear complementarity constraint (Figure \ref{fig:block_trajectories}), with mean-squared deviations less than $10^{-6}$ for state, $10^{-4}$ for control, and $10^{-3}$ for contact force trajectories. For moderate values of uncertainty ($0.01 < \sigma < 1.0$), the generated trajectories deviate from the nominal trajectory, and the magnitude of the deviation grows with the magnitude of the uncertainty. Specifically, the controls are more aggressive and nonzero for only part of the duration and the friction forces are also nonzero for only part of the trajectory. At the highest value for uncertainty we tested ($\sigma = 1.0$), the control approaches a bang-bang control, and the friction forces are zero for the duration, which indicates that the ERM may produce infeasible solutions if the uncertainty is too high.

\begin{figure}
    \centering
    \includegraphics[width=\linewidth]{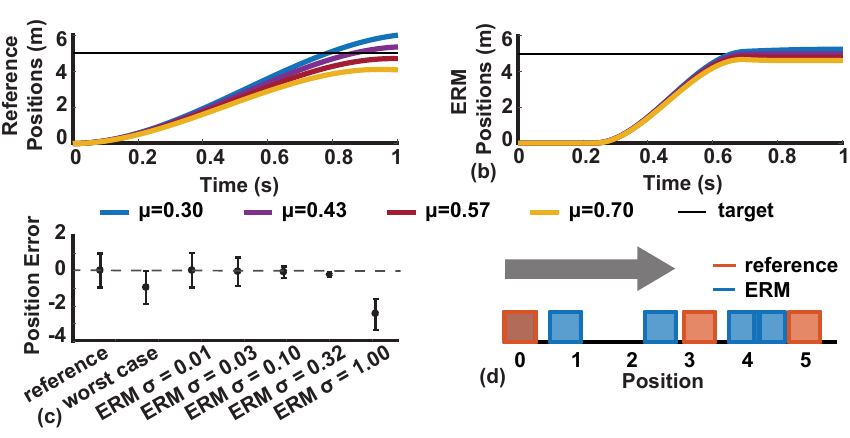}
    \caption{Comparison of trajectories generated by the reference (a) and ERM (b) controls under different values of the terrain friction. The horizontal line represents the target position. (c) The mean and range of the deviation of the simulated final position from the goal position for different models of friction. As uncertainty increases, the range of final positions under friction perturbations decreases. However, if the uncertainty is too large, the planned motion is infeasible, and the simulation produces large deviations from the desired position. (d) Selected frames of the block's motion from the reference and ERM trajectories. The arrow indicates the direction of motion.}
    \label{fig:block_simulation}
    \vspace{-5mm}
\end{figure}

In open-loop simulation, the ERM generated controls with low uncertainty ($\sigma \leq 0.01$, $\sigma < 0.01$ not shown) produced trajectories with a spread in final state similar to the reference control (Figure \ref{fig:block_simulation}). Under frictional perturbations, the ERM controls with $\sigma \leq 0.01$ resulted in final positions within 1m of the target and with an average error of 0m. As the uncertainty increased, the spread in final positions decreased from 1m to 0.22m, indicating the uncertainty produced controls that were more robust to frictional perturbations. However, when the uncertainty was $\sigma = 1.0$, the performance degraded and the open-loop average position error was -2.4m. In this case, the ERM cost landscape corresponds to that in Figure \ref{fig:ERMCostMap} ($\sigma = 10$), as the uncertainty in friction is multiplied by the normal force, $\lambda_N=9.8N$. As indicated in the figure, the ERM solution set no longer corresponds to the complementarity solution set, and therefore ERM produces a physically infeasible solution. However, this behavior is sensitive to the choice of units for the normal force; if we had instead used the normal impulse, then the uncertainty would have been multiplied by the timestep as well, and a suitable timestep could have been chosen to alter ERM cost landscape to produce feasible solutions.

In contrast, the worst-case scenario always produced a feasible trajectory with respect to at least one value of the friction coefficient in the uncertainty set. However, in open-loop simulations, the control produced by the worst-case method had an average error of $-0.95$m and a range of $-1.9$m, which was comparable to the range produced by the reference non-robust control.

\subsection{Increasing Uncertainty Increases Distance to Terrain}
\begin{figure}
    \centering
        \includegraphics[width=1\linewidth]{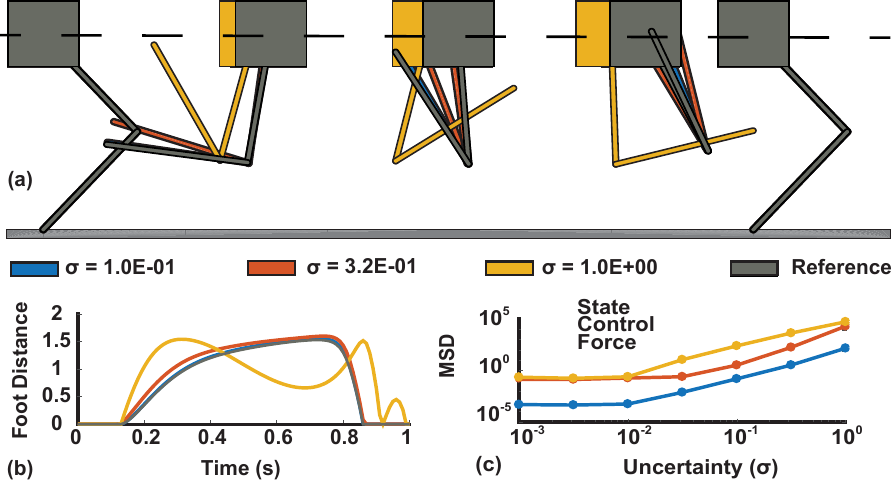}
    \caption{Illustration of the relationship between contact uncertainty and locomotion foot clearance height. As uncertainty increases, ERM increases the distance to the terrain. (a) Selected configurations of the contact-driven cart under different values of uncertainty, where the cart is constrained along a horizontal track. For $\sigma < 0.1$, the configurations are indistinguishable from the non-ERM reference trajectory. (b) The normal distance between the endpoint of the contact-driven cart and the terrain over the entire trajectory. As uncertainty increases, the distance increases until the second link flips over, decreasing the distance again. (c) Mean-squared difference between the ERM solutions and the non-ERM reference. As uncertainty decreases, the ERM trajectories converge to the reference trajectory.}
    \label{fig:ContactCart}
    \vspace{-1mm}
\end{figure}

In the contact-driven cart simulation, low values of uncertainty ($\sigma \leq 0.01$) in the ERM objective resulted in trajectories that were close to the optimal trajectory generated by complementarity constraints (Figure \ref{fig:ContactCart}). However, as the uncertainty increased, the ERM-generated trajectories increased the distance between the terrain and the end-effector. The average distance to the terrain was 0.84m  for $\sigma = 1.0$ and 0.79m for $\sigma = 0.001$, compared to an average distance of 0.79m in the reference non-robust trajectory.

\subsection{ERM Increases Foot Ground Clearance}
\begin{figure}
    \centering
    \includegraphics[width=\linewidth]{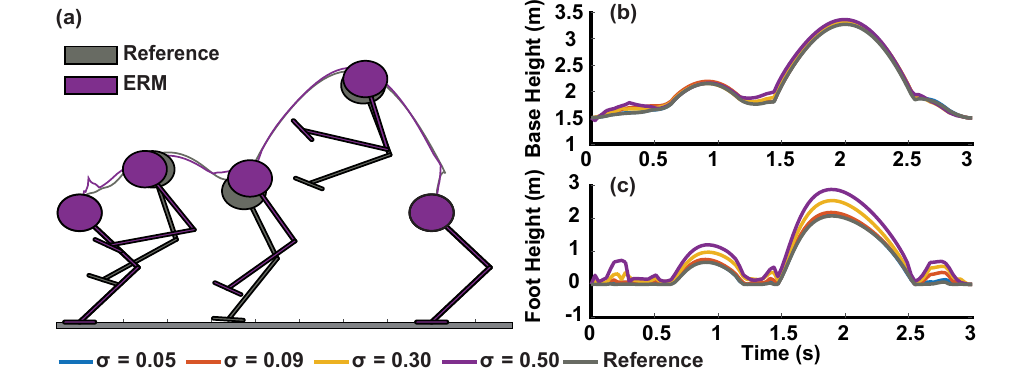}
    \caption{Illustration of the hopper experiments with both friction and terrain uncertainty. (a) Selected configurations of the hopper in both the reference trajectory and in the ERM trajectory for height uncertainty $\sigma = 0.50$ and friction coefficient uncertainty $\sigma = 0.01$. Traces indicate the center of mass trajectories. (b) Base and (c) foot height trajectories under different uncertainties. The ERM cost increases the foot, but not the base, height.}
    \label{fig:HopperJump}
    \vspace{-5mm}
\end{figure}

For all values of terrain height uncertainty tested in the hopping example, incorporating terrain and friction uncertainties increased the foot clearance of the hopper (Figure \ref{fig:HopperJump}). Moreover, the increase in foot clearance trended with the uncertainty in the terrain contact distance, with increases between 4.7\% ($\sigma = 0.05$) and 74.1\% ($\sigma = 0.50$) in our experiments (Table \ref{tab:Table 1}). In contrast, the height of the base of the hopper increases only marginally, with a minimum increase of 0.8\% for $\sigma = 0.05$ and a maximum increase of 3.1\% for $\sigma = 0.5$. In all these cases, the friction coefficient uncertainty was fixed at $\sigma_{\mu} = 0.01$.

\begin{table}
    \caption{Mean Percent Increase in Height with Height Uncertainty}
    \label{tab:Table 1}
    \centering
        \begin{tabular}{|l|c|c|c|c|}
            \hline 
            & \multicolumn{4}{c|}{Height Uncertainty} \\
            \hline 
            & $\sigma=0.05$ & $\sigma=0.09$ & $\sigma=0.30$ & $\sigma=0.50$\\
            \hline
            Foot Height & 4.7\% & 12.9\% & 43.2\% & 74.1\% \\
            \hline
            Base Height & 0.8\% & 1.6\% & 1.7\%  & 3.1\% \\
            \hline
        \end{tabular}
        \vspace{-5mm}
\end{table}
\section{DISCUSSION AND CONCLUSIONS}
Our ERM method for modeling terrain uncertainties is an important step towards a deployable terrain-robust contact-implicit trajectory optimization. One advantage of our approach over the previous work \cite{tassastochastic} is that our approach explicitly models and is robust to uncertainty in the contact parameters. By evaluating a variety of uncertainty parameters, we demonstrated that our approach generates trajectories of varying robustness and converges to the traditional, non-robust solution as the uncertainty vanishes.

The proposed ERM method is similar to the previous ensemble approach in that both achieve robustness by introducing a cost with respect to random parameter variations \cite{mordatch2015ensemble}. However, unlike the previous approach, we did not need more than one trajectory to achieve robustness. Instead, we assumed normal distributions over the friction coefficient and terrain height and calculated the expected value analytically as in \cite{tassastochastic}. A closed-form expression for the expectation allowed us to avoid sampling-based approaches. However, as in the original work \cite{tassastochastic}, we note that the ERM objective in our study has no physical meaning - the interpretation of the complementarity constraints is lost when the constraints are replaced with the residual function in Eq. \eqref{eq:LCPequivalence}. Future work may improve on our work by developing terrain-robust objectives that admit a physical interpretation.

In the context of risk-sensitive control, our approach is analogous to the risk-averse control in \cite{farshidian2015risk}. For our work under friction uncertainty, the optimization incurs little additional cost from the uncertainty if the system is at rest and there are no tangential frictional forces, provided the normal force is sufficiently large. Thus, the friction ERM cost promotes the short and fast sliding motions observed in the sliding block example. Likewise, in the uncertain terrain distance model, the ERM cost penalizes proximity to the expected terrain, and thus the system tends to move away from the terrain, using more control and taking higher steps to reach the goals.  These behaviors can be understood as risk averse, as the ERM minimizes the interactions between the system and the uncertain terrain. In contrast to our approach, risk-seeking behaviors could reward the system for making more contact interactions with the environment and could be useful for robots to collect more terrain data for estimation. 

In this work, we compared our ERM approach to uncertainty in the complementarity conditions to a worst-case solution. The worst-case solution, inspired by \cite{xie2016robust}, is a robust, distribution-free method to solve complementarity problems such as the contact conditions in contact-implicit trajectory optimization. However, unlike our ERM method, the worst-case method assumes a discrete set of uncertain values and finds the solution for the value in the set that maximizes the complementarity residual. This is analogous to choosing a particular value of the uncertain parameter (for example, choosing the friction coefficient), and then solving the corresponding optimization. Here, we have shown that, while the worst-case may achieve a robust solution to the complementarity problem, that robustness does not translate to the generated controls, as the worst-case solutions produce open-loop trajectories with the same endpoint variation as the standard contact-implicit method. Thus, although our approach may not strictly satisfy the complementarity constraints for all values of the uncertain parameters, it does have an advantage over the worst-case method in that the control trajectories inherit robustness from the ERM solutions.

One important feature of our work is that, as the uncertainty approaches zero, the trajectories approach the solutions generated by using the mean value of the uncertain parameters. While the low uncertainty case can be interpreted as a smooth and accurate approximation to the original nonsmooth complementarity constraint, we also note that the property alone is important, as it opens an avenue for combining model-based approaches, such as contact-implicit trajectory optimization, with model-free Bayesian optimization methods \cite{kuindersma2013variable, seyde_locomotion_2019}. Future work could combine our work here with measurements from the terrain to estimate the terrain parameters during locomotion and close the loop of terrain estimation and robust trajectory optimization.

\vspace{0.2in}

\bibliographystyle{IEEEtran}
\bibliography{lidar.bib}
\end{document}